# What has ChatGPT read? The origins of archaeological citations used by a generative artificial intelligence application

Dirk HR Spennemann

*School of Agricultural, Environmental and Veterinary Sciences; Charles Sturt University; PO Box 789; Albury NSW 2640, Australia.*

*email: dspennemann@csu.edu.au*

*ORCID: 0000-0003-2639-7950*

## Abstract

The public release of ChatGPT has resulted in considerable publicity and has led to wide-spread discussion of the usefulness and capabilities of generative AI language models. Its ability to extract and summarise data from textual sources and present them as human-like contextual responses makes it an eminently suitable tool to answer questions users might ask. This paper tested what archaeological literature appears to have been included in ChatGPT's training phase. While ChatGPT offered seemingly pertinent references, a large percentage proved to be fictitious. Using cloze analysis to make inferences on the sources 'memorised' by a generative AI model, this paper was unable to prove that ChatGPT had access to the full texts of the genuine references. It can be shown that all references provided by ChatGPT that were found to be genuine have also been cited on Wikipedia pages. This strongly indicates that the source base for at least some of the data is found in those pages. The implications of this in relation to data quality are discussed.

## Introduction

The application of artificial intelligence (AI) in numerous fields of research and professional endeavour has gained widespread public notice in recent months. The public release of DALL-E (an image generator) and, in particular, of the Chat Generative Pre-trained Transformer (ChatGPT) in early 2022 captured the public imagination, sparking a wide- and free-ranging discussion not just on its current and prospective future capabilities, but also on the dangers this may represent, as well as the ethics of its use. ChatGPT is an OpenAI generative AI language model that leverages transformer architecture to generate coherent and contextually appropriate, human-like responses based on the input it receives (OpenAI 2023).

ChatGPT has gone through several iterations and improvements since its formal release in 2018, with the focus placed on increased text prediction capabilities, including the ability to provide longer segments of coherent text and the addition of human preferences and feedback. ChatGPT 2.0, released in September 2019, drew on a training data set that relied on 1.5 billion parameters. ChatGPT 3 (released in June 2020) was trained (by humans) on 175 billion parameters. The key advancement was its ability to perform diverse natural language tasks, such as text classification and sentiment analysis, facilitated contextual answering of questions, allowing ChatGPT to function not only as a chatbot, but also to draft basic contextual texts like e-mails and programming code.

The first version accessible to the general public, ChatGPT 3.5, was released in November 2022 to the general public, as a part of a free research preview to encourage experimentation (Ray 2023). The current version GPT-4, released in March 2023, reputedly exhibits greater factual accuracy, reduced probability of generating offensive or dangerous output and greater responsiveness to user intentions as expressed in the questions/ query tasks. The temporal cut off for the addition of training data for ChatGPT 3.5 was September 2021, which implies that ChatGPT cannot integrate or comment on events, discoveries and viewpoints that are later than that date. It has been asserted in public media, however, that GPT-4 has the ability to search the Internet in real time.

Given the popularity, there is a growing body of research that investigates the capabilities and level of knowledge of ChatGPT and its responses when queried about numerous fields of research. Several papers have examined ChatGPT's understanding of disciplines such as agriculture (Biswas 2023), chemistry (Castro Nascimento and Pimentel 2023), computer programming (Surameery and Shakor 2023), cultural heritage management (Spennemann 2023a, in press) , diabetes education (Sng et al. 2023), nursing education (Qi, Zhu, and Wu 2023), and remote sensing in archaeology (Agapiou and Lysandrou 2023). Examinations in the various fields of medicine are particularly abundant (King 2023; Sarraju et al. 2023; Bays et al. 2023; Grünebaum et al. 2023; Rao et al. 2023).

In the cultural heritage field, ChatGPT's abilities have been examined in museum settings, in particular in terms of its ability to provide visitor guidance (Trichopoulos, Konstantakis, Alexandridis, et al. 2023) and to develop exhibition texts, exhibit labels and catalogue information, as well as scripts for audio guides (Maas 2023; Merritt 2023; Trichopoulos, Konstantakis, Caridakis, et al. 2023). Other work examined its ability to assist curators in developing exhibition concepts (Spennemann in press). Outside museum studies, the use of ChatGPT has seen only little attention, with one paper examining its ability to explain the use of remote sensing in archaeology (Agapiou and Lysandrou 2023) and another looking into its understanding of value concepts in cultural heritage (Spennemann 2023a).

At present, ChatGPT has the ability to answer questions and general queries about topics by summarising information contained in the training data. The human-conversation style interface, allowing for ongoing 'conversations,' gives a user the opportunity to pose follow-up questions asking for expanded and deepened information. Thus, at least at first sight, ChatGPT is poised to become an eminently suitable tool for public interaction and public education in heritage/historic preservation or public archaeology.

ChatGPT often purports to merely strive to provide factual and neutral information and not to hold political opinions (Rozado 2023). Yet, as several authors have pointed out, ChatGPT cannot be without bias as model specifications, algorithmic constraints and policy decisions shape the final product (Ferrara 2023). Moreover, the source material that was entered into its dataset during its training phase, was in turn subconsciously or consciously, influenced if not shaped by the ideologies of the people programming and 'feeding' the system. Consequently, political orientation tests showed that ChatGPT is biased and ChatGPT exhibits a preference for libertarian, progressive, and left-leaning viewpoints (Rozado 2023; Rutinowski et al. 2023;

Motoki, Pinho Neto, and Rodrigues 2023; McGee 2023; Hartmann, Schwenzow, and Witte 2023), with a , North American slant (Cao et al. 2023).

While ChatGPT as a generative language model is generally good at collating, extracting and summarising information that it was exposed to in its training data set, its accuracy is based on statistical models of associations during training and its frequency (Elazar et al. 2022). It lacks reasoning ability (Bang et al. 2023) and is thus unable to provide essay-based assignments, let alone academic manuscripts, of an acceptable standard (Fergus, Botha, and Ostovar 2023; Hill-Yardin et al. 2023; Wen and Wang 2023). Moreover, ChatGPT has also been shown to, at least occasionally, suffer from inverted logic (Spennemann 2023a), ultimately providing disinformation to the reader. When asked to provide academic references to its output of assignments or essays, ChatGPT is known to 'hallucinate' or 'confabulate' (Millidge 2023; Bang et al. 2023; Alkaissi and McFarlane 2023), generating both genuine and spurious output (Athaluri et al. 2023; Spennemann 2023a). While the researchers working with and examining the capabilities of ChatGPT are aware of this, the general public, by and large, is not.

Given the potential of ChatGPT to become an eminently suitable tool for public interaction in public archaeology/heritage, it is of interest to understand, what ChatGPT 'knows' about archaeology, historic preservation and cultural heritage management in general. Clearly, AI reasoning aside, any collated and synthesised information provided by ChatGPT can only be as good as the data it has access to. This raises the question, what data were made available to ChatGPT during its training phase, as full text, as partial text ('snippets') and which sources ChatGPT was made aware of without access to full or partial text. This is critical as, biases aside, any aggregation of information by ChatGPT can only be as good as the sources it can draw on. At present only one such study exists, which examined the access and utilisation of general literature. Using cloze analysis, which uses text snippets with missing text to make inferences on the sources 'memorised' by a generative AI model (Shokri et al. 2017; Tirumala et al. 2022; Onishi et al. 2016), Chang et al. (2023) found that the "degree of memorization was tied to the frequency with which passages of those books appear on the web".

This paper will examine what archaeological literature appears to have been made available to ChatGPT and included in its training phase, based on self-reporting of references. All references were assessed as to their veracity and the origin and nature of the genuine references will be discussed.

## Methodology

The study used OpenAI's ChatGPT (https://chat.openai.com [for access dates and versions see Table 1]. ChatGPT was tasked with the following request: "Cite [number] references on [topic]" where the requested number was either 20 or 50. The topics were 'cultural values in cultural heritage management' (building on an initial analysis in Spennemann 2023a), 'archaeological theory', 'Pacific archaeology' and 'Australian archaeology' (see Table 1). As ChatGPT can draw on prior conversations within a chat, each chat was deleted at the completion of each run, thus clearing its history. Some responses required user interaction when being prompted to "continue generating".

Runs R1 and R3 tasked ChatGPT to cite "20 references on cultural values in cultural heritage management." Once these were delivered ChatGPT was tasked with the follow up request "can you cite 20 more?" Runs R2 and R4 to R7 tasked ChatGPT to cite 50 references on a given topic. Once delivered, ChatGPT was asked to regenerate its response using the provided option. Regeneration occurred once for R2 and R4 and twice for R5 to R7, resulting in a theoretical maximum of 150 references (incl. duplicates) for each topic. The actual number of references provided is less as ChatGPT baulked at some requests due to server-demand issues (see 'Results'). Runs R1 and R2 were carried out on 17 June 2023 (version March 2023) as part of another paper (Spennemann 2023a), while runs R3 to R7 were carried out for this paper on 27 July 2023 (version July 2023).

The veracity of all references was ascertained through title searches in GoogleScholar. The accuracy of each reference was assessed in terms of 'author(s),' 'title,' 'year' and publisher/journal.' Each reference was classified as 'correct', 'correct but incorrect year,' 'confabulated,' and 'acknowledged as fictional.' To examine whether ChatGPT had access to a full text, or only the free preview sections of Google Books, for example, the cloze prompt methodology used by Chang et al. (2023) was utilised. In this, ChatGPT is tasked with the identification of a proper noun missing in a sentence taken from an original text that is suspected to have been used as training data (see Appendix F for script). A set of 10 sample sentences were used for each source, with five sentences drawn from text accessible via Google Books Preview, and five sentences for which access to an actual copy of the work was required. Chat GPT (August 3 [2023] version) was provided with each sentence in turn and asked to regenerate each answer. After the regenerate prompt, ChatGPT was given no indication whether the second answer was better or not. The whole sequence was then repeated in the same chat session. The chat was then closed and deleted. The whole process was repeated with the same sample sentences in a new chat session. The sample sentences and the word identifications are reported in Appendices G to I).

All conversations with ChatGPT used in this paper have been documented according to a protocol (Spennemann 2023b) and have been archived as a supplementary data file at XYZ [to be inserted upon publication].

## Results and Discussion

Throughout, ChatGPT responded in a helpful tone to the requests and the responses are phrased in qualifying terms: "As an AI language model, I don't have direct access to databases or external sources, and I can't generate citations in a conventional academic format" (run 3, set 1); or: "As an AI language model, I don't have direct access to databases or external sources such as specific references or papers"(run 7, set 2). Despite these qualifiers, ChatGPT then proceeded to offer lists of references, indicating that they were both valid and pertinent: "However, I can suggest a diverse list of 50 references on Australian archaeology that you can explore further"(run 7, set 2); or "However, I can provide you with a list of key archaeological theorists and their seminal works that you can use as references for your research" (run 5 set 1). It usually concludes the reference list provided with comments like "Remember to follow the specific citation style (e.g., APA, MLA, Chicago) required by your academic institution or publication when using these references" (run 5, set 1). On occasions it suggests to "Make sure to verify the relevance and quality of each reference for your specific research purposes"(run 7, set 2).

### *Authenticity of References*

To a casual user who is not familiar with the literature, *all* references appear valid because i) the titles appear plausible; ii) journal titles are those of genuine publications and iii) and the vast majority of author names cited were those of academics actively publishing in the fields of cultural heritage or archaeology. With one exception (of 17 sets of reference lists) there is no indication that references provided by ChatGPT might not have been genuine. When asked to generate 50 sample references related to cultural values in cultural heritage management, it provided the following caveat before offering the references: "Please note that some of these references might be fictional, and it's essential to verify their accuracy and relevance in academic and research contexts" (run 4 set 2).

The propensity of ChatGPT to generate fictitious references has been noted before (Athaluri et al. 2023; Spennemann 2023a; Day 2023; Gravel, D'Amours-Gravel, and Osmanlliu 2023). An examination of the non-existent references shows that these were generated using the names of real authors working in the field (in the main) with fragments of real article titles, journal or publisher names to construct false but realistic looking references. An example is the following reference (run 6 set1): Best, S., & Clark, G. (2008). Post-Spanish Contact Archaeology of Guahan (Guam). Micronesian Journal of the Humanities and Social Sciences, 7(2), 37-74. This reference can be deconstructed as follows:

| | |
|---|---|
| Best, S., | existing author, Simon Best |
| Clark, G. | existing author, Geoff Clark |
| (2008). | plausible year |
| Post-Spanish Contact Archaeology | contextually plausible time frame in title |
| of Guahan (Guam). | plausible location |
| Micronesian Journal of the Humanities and Social Sciences | journal on record |
| 7(2), | does not exist, journal ceased with volume 5, 2006 |
| 37-74. | irrelevant as volume incorrect |

A second example is the following reference (R5 set 1): "Bintliff, J. L. (1991). The Annales School and Archaeology. In Theoretical Roman Archaeology: Second Conference Proceedings (pp. 61-84). Oxbow Books." This reference can be deconstructed as follows:

| | |
|---|---|
| Bintliff, J. L | correct author for first part of title, John L. Bintliff |
| (1991). | correct year for first part of title |
| The Annales School and Archaeology. | existing book not chapter title (Bintliff 1991) |
| Theoretical Roman Archaeology: Second Conference Proceedings | existing book title, but (Rush 1995) |
| (pp. 61-84). | pagination not connected with Rush |
| Oxbow Books | existing publisher, but unrelated to either part of the title |

In this case, both components were listed in incomplete form on the site philapers.org (https://philpapers.org/rec/RUSTRA-3 and https://philpapers.org/rec/BINTAS), possibly pointing to the source material. A variation of the above confabulated example is the following reference (run 5 set 3): "Bintliff, J. (1991). The Annales School and Archaeology. In T. C. Champion (Ed.), Centre and Periphery: Comparative Studies in Archaeology (pp. 36-47). Unwin Hyman."

Other confabulated references are truncated versions of genuine references with spurious text. The reference "Conkey, M. W., & Tringham, R. (Eds.). (1995). Archaeology and the goddess: Exploring the contours of feminist archaeology. University of California Press" (run 3 set 1) carries the correct names, year and title, but lacks the fact that it was published as a book (not edited by the authors as suggested by ChatGPT) by a different publisher: "Margaret Conkey and Ruth Tringham (1995) Archaeology and The Goddess: Exploring the Contours of Feminist Archaeology. In *Feminisms in the Academy: Rethinking the Disciplines*, edited by A. Stewart and D. Stanton, pp. 199-247. University of Michigan Press, Ann Arbor."

Another example of a near genuine yet confabulated reference is the following (run 3 set 2): "DeSilvey, C., & Edensor, T. (2012). Reckoning with ruins. Progress in Human Geography, 36(4), 475-507." Correctly cited are the authors, and the titles of the article and the journal (DeSilvey and Edensor 2013). Incorrect are the year of publication (2012 instead of 2013) and the volume (36 instead of 37) as well as the pagination. Intriguingly, the volume/year combination of the confabulated reference (2012 v 36) is correctly constructed and tallies with the volume/year combination of the actual journal. The pagination of the confabulated reference, however, seems extracted from vol 27(4) 475-507.

Entirely fictious, but not identified as such, is the following reference with a rather humorous title offered by ChatGPT (run 4 set 1): "Van der Aa, B., & Timmermans, W. (2014). The Future of Heritage as Clumsy Knowledge. In The Future of Heritage as Clumsy Knowledge (pp. 1-13). Springer, Cham." In this instance only the publisher and the authors are genuine (Bart J.M. van der Aa, having published in the field of cultural heritage and Wim Timmermans, having published in urban landscapes).

## *Authenticity misrepresented*

Having highlighted a high percentage of confabulated references, the question arises as to how ChatGPT represents the references it provided. In two of the responses ChatGPT actually offers the user a "*curated* list of important and influential references in the field of archaeological theory" (run 5 set 2) or a "*curated* list of 10 reputable references on Australian archaeology" (run 7 set 1). The choice of the word 'curated', which is heavily overused these days (Edmundson 2015), insinuates that the user can expect a set of the references that have been carefully selected to best represent the literature on the topic. Of the fifteen references on archaeological theory, twelve (80%) are correct and three are confabulated. While better than the overall average (68.7%, Table 2), this does not constitute 'curated.' In its 'curated' list of references on Australian archaeology, the percentages are reversed, with eight of the ten references confabulated. Moreover, the remaining two are cited with an incorrect year. In another instance, ChatGPT claimed that it could "certainly provide you with a list of 10 reputable references on cultural values in cultural heritage management" (run 2 set 1). Six of these were confabulated while four were correct. Clearly, any claim of 'curated' or 'reputable' is completely misleading.

## *The sources of the genuine references*

The potential origin of those references that had been deemed genuine (Appendices A–D) was assessed by examining whether they were accessible via Google Books or other sources (Table 3). The overwhelming majority of references existed in Google Books (82.5%). In terms of access, the full text of 16.8% of all references could be freely accessed (via Google Books, JSTOR or otherwise online), while close to two thirds were (66.4%) accessible in Google Books preview mode, which shows some, but not all of the text. The text of 10.7% of all cited genuine references was not accessible online, which would require them to be manually 'borrowed' (via Archive.org) (Table 3).

In a different context, ChatGPT's knowledge regarding the emerging heritage of COVID-19 was critically examined. This showed that some material must have come either from the media or from Wikipedia. The latter seemed the most likely {Spennemann, in press #24206}. To test whether all genuine sources were also cited on Wikipedia pages, each genuine reference was checked in Google, using the following search logic: "citation" + site:Wikipedia.org. The individual results are shown in Appendices A–D. It appears that *all* genuine sources have been cited in Wikipedia or its sibling such as Wikidata. Moreover, the higher percentage of genuine citations in the fields of archaeological theory compared to the other three disciplines assessed (Table 2), reflects the greater volume of Wikipedia text dedicated to that discipline and the biographies of archaeological theorists, compared to the other disciplines.

This then raises the question whether the 'knowledge' used by ChatGPT stems from primary sources, as the genuine citations might indicate, or whether it is based on the secondary source texts of the Wikipedia pages.

### *Did ChatGPT have access to the full texts of sources?*

To examine whether ChatGPT had access to a full text, or only to the free preview sections of Google Books, for example, the cloze prompt methodology used by Chang et al. (2023) was utilised. Tested were three sources: i) Bruce Pascoe's *Dark Emu* (Appendix G)(Pascoe 2014); ii) Alison Wylie's edited book Thinking from Things (Appendix H)(Wylie 2002) ; and iii) Patrick Kirch and Roger Green's 'Hawaiki, Ancestral Polynesia' (Appendix I) (Kirch and Green 2001).

In *Dark Emu* only one sample sentence (6) was consistently correctly answered by ChatGPT by providing the missing word 'Glock.' In sample sentence 10 it correctly provided the missing word 'Stanner,' in all but the first iteration (Appendix G2 and G3). Both sample sentences stem from sections of the book that are not publicly available via Google Preview. ChatGPT provided incorrect answers in the first response set (initial answer and regenerated answer) in both runs 1 and 2 but provided correct responses ('Gammage') in the second response set (Appendix G2 and G3). All other answers supplied by ChatGPT are incorrect. For sample sentence 10, ChatGPT offered incorrect responses throughout run 1 and the initial answer of the first response set in run 2, but then provided correct responses ('Sturt') for the remainder.

In *Thinking from Things* two sample sentences (6 and 7) were consistently correctly completed by ChatGPT by providing the missing words 'Laudan' and 'Sutton' respectively. Both sample sentences stem from sections of the book that are not publicly available via Google Preview. Among the other eight sample sentences, ChatGPT provided a single correct answer for sentences 1 and 2 (both in the first response set) (Appendix H2 and H3).

Half the sample sentences of the last sample set, taken from *Hawaiki, Ancestral Polynesia*, were consistently correctly answered (1,2 6, 7, 9), while sample sentence 8 was correctly answered in seven of the eight responses. The incorrect answer occurred during the regeneration of the initial answer of the run 2 (Appendix I2 and I3). ChatGPT was able to consistently provide correct answers to sample sentence 10 in run 1 but provided incorrect answers throughout run 2.

In comparison, the percentage of correct substitutions for 'Dark Emu' and 'Thinking from Things' was 40% or less in all cases, with no discernible prevalence of publicly accessible text over non-accessible sections (Table 4). While the difference in the success rate of correct answers between 'Dark Emu' and 'Thinking from Things' is minimal and insignificant (paired T-TEST, $p$=0.275) the success rate of correct answers for 'Hawaiki, Ancestral Polynesia' was significantly higher than any of the other two sources (paired T-TEST, $p < 0.002$ Emu, $p < 0.001$ Things).

### *Did ChatGPT actually access the full texts of sources?*

While the majority of the genuine references could be tracked back to possible sources (Table 3) the question arises whether ChatGPT actually 'read' these sources during its training phase, or whether it drew those references from pre-compiled lists or obtained them from other sources.

There were eleven instances, where ChatGPT consistently provided the correct answer to the cloze prompt question. To test, whether publicly accessible sources other than Google Books may have been used as training data, the respective phrases were entered as full text searches into Google (confined by quotation marks as a phrase). In the case of 'Dark Emu,' the full text search for the sample sentence 6 found two instances, Google Books and Weibo, while the search for the sample sentence 9 found copies in Google Books and on the site VDOC.pub. In the case of 'Thinking from Things,' a full text search for the sample sentence 6 found two instances, Google Books and Scribd, while a search for sample sentence 7 only returned Google Books. The largest number of consistently correct answers was encountered in 'Hawaiki, Ancestral Polynesia,' with sample sentences 2, 7 and 10 located in Google Books and a file uploaded to DocPlayer, sample sentences 6 and 8 accessed via Google Books and sample sentence 9 accessed via a file uploaded to DocPlayer. Sample sentence 1, however, could not be located at all. Of note in this regard is the observation that while sample sentences 6 and 10 of *Dark Emu* may not be publicly accessible via Google Books, they can still be found via a simple Google search where the phrase text is linked back to the correct section in Google Books (where it is blocked from view, however). This indicates that the full text of a publication is included in Google's database even if only a section is publicly visible.

As noted earlier, ChatGPT exhibits a tendency to confabulation, as evidenced by the pattern observed among the fictional references that all author names appeared plausible because they were published anthropologists or archaeologists and thus fitted the context. Given that ChatGPT is a generative language model that uses transformer architecture to generate coherent and contextually relevant responses, it is therefore possible that cloze analysis does not prove actual retrieval of information from the training data set, but that, due to the extent of specific context and conceptual detail provided, some text samples may be more likely to result in correct answers. In *Dark Emu,* for example, only one sample sentence (6) was consistently correctly completed, which, upon reflection, provided specific context and conceptual detail: "So the police escorted the 4WD heroes into the initiation site. Once there, they threw beer cans into the sacred water, and took it in turns to shoot at the cans with police-issue Glock pistols", ChatGPT correctly completed the missing word 'Glock' (Appendix G2 and G3). Given the context of 'shoot at the cans' and 'police-issue … pistols', and a setting in Australia, this considerably narrows the substitution of the missing word to 'Glock.'

To test whether specific context and conceptual detail would influence ChatGPT responses, sample sentences 6, 7 and 9 taken from '*Hawaiki, Ancestral Polynesia*' were retested, with different proper nouns omitted. One initial answer and three regenerations in the same chat were sought. Sample sentence 6 listed three anthropologists in succession: "The founders of the unique Americanist tradition in anthropology - Boas, Kroeber, Sapir, and others - reacted in part to the theoretical excesses …" The initial assessment asked ChatGPT to identify 'Kroeber' as the missing word, which it did consistently. When asked to find the correct replacement for the missing first name, Boas, it also did so consistently, answering either 'Franz Boas' or 'Boas.' When asked to find the correct replacement for the first name, Sapir, however, it offered three different incorrect names, Lowie, Malinowski, and Mead, with Malinowski offered again in the last regeneration. In sample sentence 7, which starts with "In reading Flannery and Marcus otherwise brilliantly argued volume…", ChatGPT had to find the correct replacement for the missing name 'Marcus' which it did consistently. When tasked with finding the correct replacement for 'Flannery', however, it offered three incorrect names, Trudgill, Lyle, Greenhill, with Greenhill repeated. The final assessment used sample sentence 9, which in its core contains the section "…to paraphrase the great evolutionist George Gaylord Simpson, "one cannot…" ChatGPT had to find the correct replacement for the missing name 'Gaylord', which it did consistently. The same applied when the missing name was 'George.' Once the final name in the sequence was the target of identification, however, ChatGPT, offered 'Darwin' three times in succession and finally answered correctly with 'Simpson.'

These examples indicate that ChatGPT does not seem to be able to 'reach back' into its training dataset and to extract text quotes. There are two possible explanations. Either cloze analysis is unsuitable as a method for source identification seemingly contradicting Chang et al. (2023), or the training data set for ChatGPT did not include the texts of these genuine references. Given that the study by Chang et al. (2023) appears convincing, the inability to prove that ChatGPT actually 'read' primary sources confirms earlier doubts and suggests that much of its knowledge base is rooted in the secondary material contained in the Wikipedia universe.

## Conclusions and Implications

The public release of ChatGPT has resulted in considerable publicity and has led to wide-spread discussion of the usefulness and capabilities of generative AI language models. Its ability to extract and summarise data from textual sources and present them as human-like contextual responses makes it an eminently suitable tool to answer questions users might ask in museum settings or for public interaction and public education in heritage/historic preservation or public archaeology. Any collated and synthesised information provided by ChatGPT, however, can only ever be as good as the data it has access to.

A museum, or similar applied setting (e.g., public interpretation of an excavation site), provides a closed system, where the quality and authenticity of data can be ensured. In a museum setting, for example, high quality training data can be provided in the form of information on the purpose and mission of the museum, exhibition content and layout, as well as details on all items in the collection, their acquisition history and the cultural or natural history contexts in which they are situated. Adequate quality control by human trainers during the training phase can ensure that erroneous connections and confabulations are being minimised.

Beyond closed systems, however, ChatGPT has the propensity to generate text based on associations, where some generated text may at first sight appear plausible, but which upon closer examination is found to be flawed or even wrong. This paper tested what archaeological literature appears to have been made available to ChatGPT and included in its training phase. While ChatGPT could readily provide seemingly pertinent references, a large percentage were fictitious. To a casual user who is not familiar with the literature, *all* references appear valid, however, because the titles were plausible and the vast majority of authors cited were those of academics actively publishing in the fields of cultural heritage or archaeology.

As no information on the detailed nature of training data has been made public, we do not know which, if any, texts of the genuine references have been included in the training data set. The author was unable to prove that ChatGPT accessed any of the actual texts of the genuine references in its training phase. An examination has shown, however, that all references cited by ChatGPT that were found to be genuine have also been cited on Wikipedia pages. This strongly indicates that the source base for at least some of the data is found in those pages.

As noted above, any collated and synthesised information provided by ChatGPT, can only ever be as good as the data it has access to. Caveat emptor!

Table. 1. Experimental parameters and returned numbers of references for each set (in brackets)

| Run | Date/Time (GMT) | Version | # | Topic | Set 1 | Set 2 | Set 3 |
|---|---|---|---|---|---|---|---|
| R1 | 17-Jun-23 04:35 | May-24 2023 | 20 | cultural values in CHM | Response (20) | 20 more (2) | — |
| R2 | 17-Jun-23 04:43 | May-24 2023 | 50 | cultural values in CHM | Response (10) | 30 more (20) | — |
| R3 | 27-Jul-23 07:45 | July 2023 | 20 | cultural values in CHM | Response (20) | 20 more (20) | — |
| R4 | 27-Jul-23 07:48 | July 2023 | 50 | cultural values in CHM | Response (50) | Regenerated (50) | — |
| R5 | 27-Jul-23 07:49 | July 2023 | 50 | archaeological theory | Response (50) | Regenerated (15) | Regenerated (50) |
| R6 | 27-Jul-23 07:57 | July 2023 | 50 | Pacific archaeology | Response (50) | Regenerated (12 + 30) | Regenerated (45) |
| R7 | 27-Jul-23 21:33 | July 2023 | 50 | Australian archaeology | Response (10) | Regenerated (50) | Regenerated (50) |

Table 2. Authenticity of References

| | correct citation | wrong year cited | confabulated citation | acknowledged as fictional | n |
|---|---|---|---|---|---|
| Archaeological Theory | 68.7 | 2.6 | 28.7 | — | 115 |
| Cultural Heritage Management | 26.2 | 15.2 | 34.8 | 23.8 | 110 |
| Pacific Archaeology | 27.2 | 6.4 | 66.4 | — | 125 |
| Australian Archaeology | 3.6 | 11.8 | 84.5 | — | 210 |
| All sources | 32.1 | 10.2 | 48.8 | 8.9 | 560 |

Table 3 Possible sources of the genuine references

| | Arch. Theory | Australian Arch. | CHM | Pacific Arch. | All |
|---|---|---|---|---|---|
| Google Books, preview with sections | 64.5 | 50.0 | 74.4 | 61.5 | 66.4 |
| Google Books, no preview | 8.1 | 50.0 | 5.1 | 15.4 | 9.9 |
| JSTOR | 19.4 | 0.0 | 2.6 | 0.0 | 9.9 |
| Full text otherwise accessible online | 1.6 | 0.0 | 10.3 | 0.0 | 3.8 |
| Google Books, full text | 0.0 | 0.0 | 0.0 | 15.4 | 3.1 |
| Google Books, snippet view | 0.0 | 0.0 | 7.7 | 7.7 | 3.1 |
| Archive.org full text | 4.8 | 0.0 | 0.0 | 0.0 | 2.3 |
| Archive.org online borrowable | 1.6 | 0.0 | 0.0 | 0.0 | 0.8 |
| n | 62 | 4 | 39 | 26 | 131 |

Table 4 Percentage of correct identifications

| Run | Set | Answer | Dark Emu | Thinking from Things | Hawaiki |
|---|---|---|---|---|---|
| 1 | 1 | Initial | 10 | 20 | 70 |
| | | Regenerated | 20 | 30 | 70 |
| | 2 | Initial | 30 | 30 | 70 |
| | | Regenerated | 30 | 20 | 70 |
| 2 | 1 | Initial | 20 | 20 | 60 |
| | | Regenerated | 30 | 20 | 40 |
| | 2 | Initial | 40 | 20 | 50 |
| | | Regenerated | 40 | 20 | 50 |
| | | Average | 27.5 | 22.5 | 60.0 |

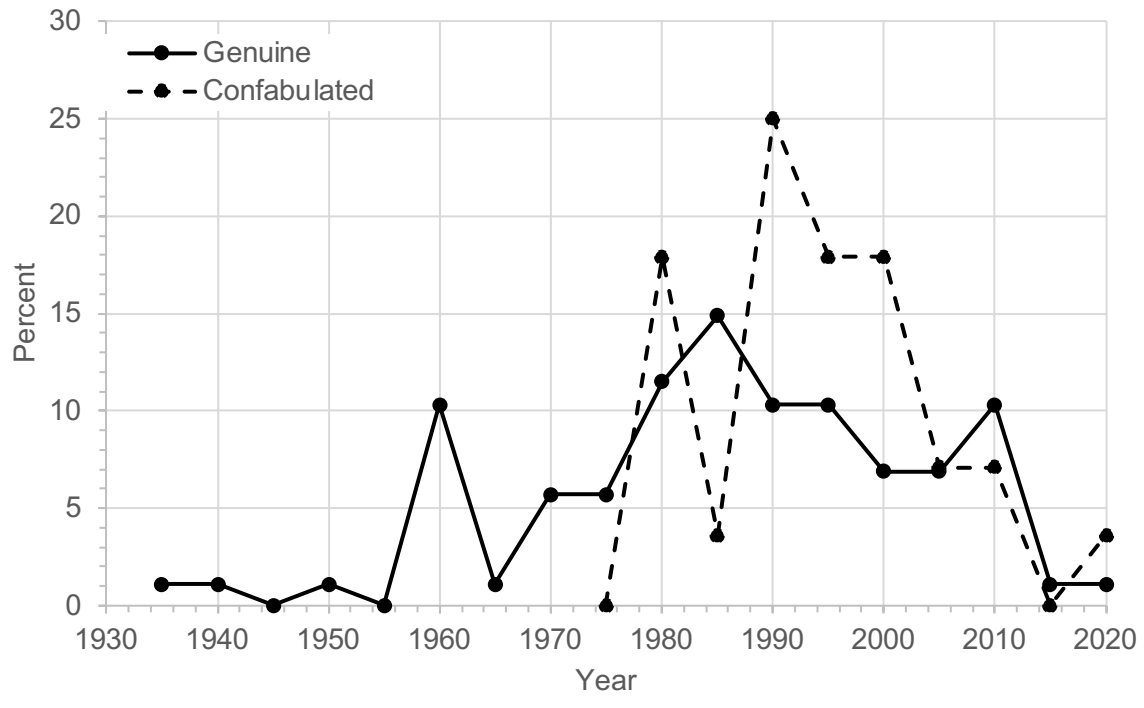

a) Archaeological Theory

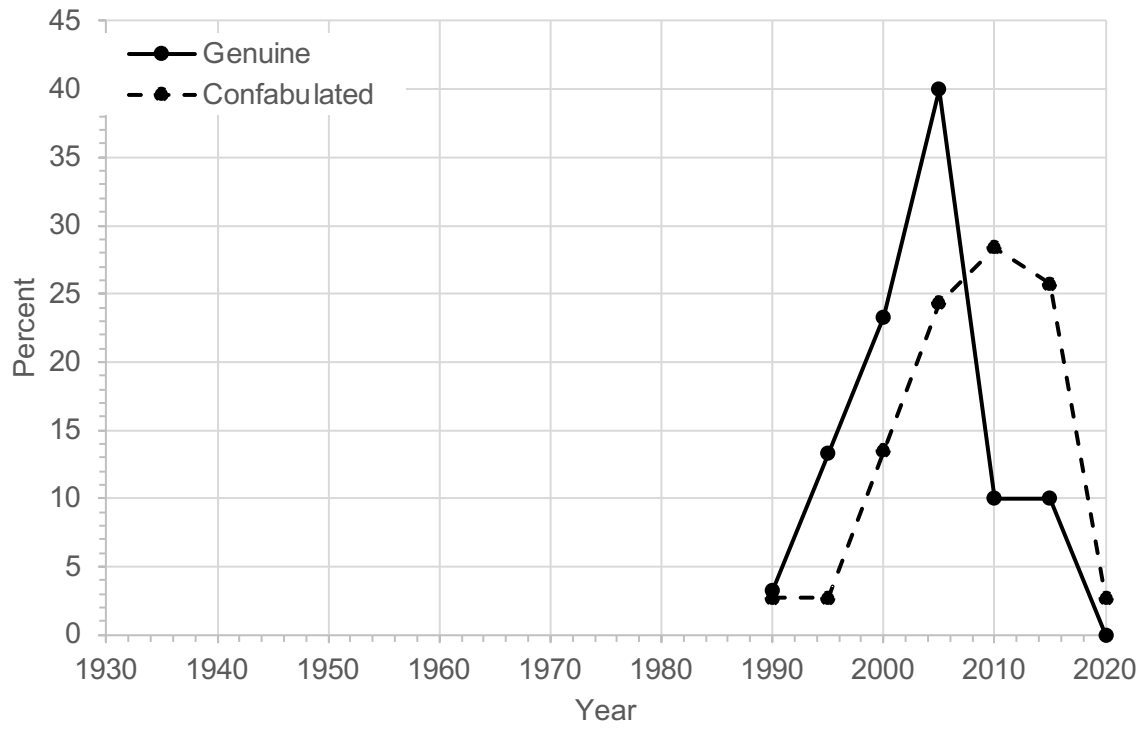

b) Pacific Archaeology

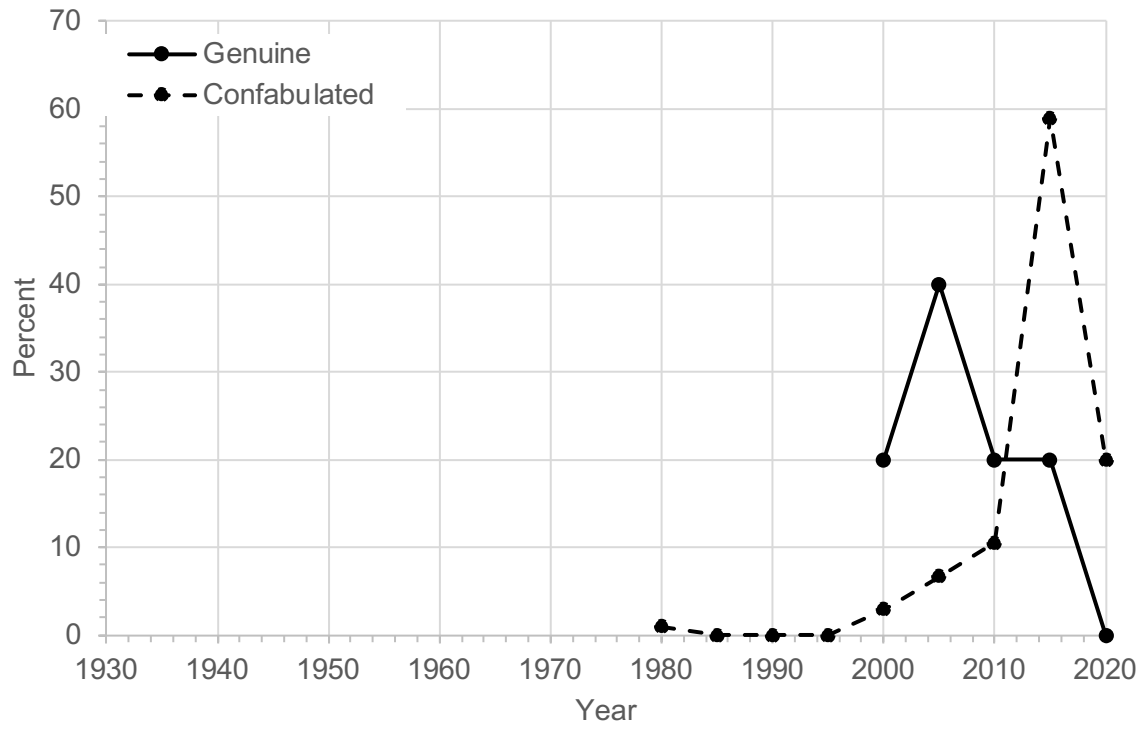

c) Australian Archaeology

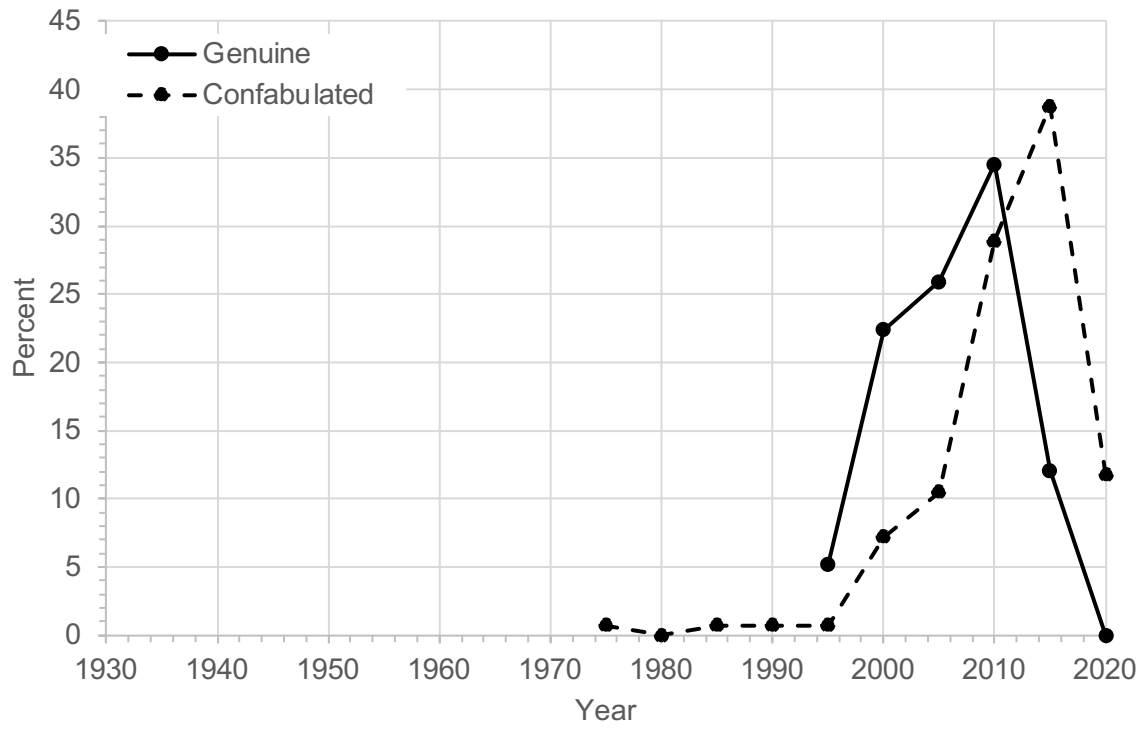

d) Cultural Heritage Management

Figure 1. Year of publishing as attributed to genuine vs confabulated and fictitious references (in five-year cohorts; for data see Appendix E).